\documentclass[11pt]{article}

\usepackage[preprint]{acl}

\usepackage{times}
\usepackage{latexsym}
\usepackage[T1]{fontenc}
\usepackage[utf8]{inputenc}
\usepackage{microtype}

\usepackage{booktabs}
\usepackage{multirow}
\usepackage{array}
\usepackage{amsmath}
\usepackage{amssymb}
\usepackage{graphicx}
\usepackage{xcolor}
\usepackage{tikz}
\usepackage{pgfplots}
\pgfplotsset{compat=1.18}
\usetikzlibrary{patterns}

\title{In-Context Fixation: When Demonstrated Labels Override Semantics
in Few-Shot Classification}

\author{Ming Liu \\
  Amazon \\
  \texttt{mlliuz@amazon.com}}

\begin{document}
\maketitle

\begin{abstract}
While random demonstration labels barely hurt in-context learning
\citep{min2022}, we show that \emph{homogeneous} labels---even
semantically valid ones---collapse accuracy to $\leq$12\% across six
models (Pythia, Llama, Qwen; 0.8B--8B) and four tasks.  The trigger is
\emph{label-slot content}: the model treats tokens occupying the label
position as an exhaustive answer vocabulary, with homogeneity as the
maximally collapsed case.  A novel \emph{set-level fixation} finding
confirms this: when demonstrations carry \emph{varied}
nonsense tokens from \{foo,bar,vex,nit,orb\}, the model places 42--67\% of
probability on the demonstrated set while $P(\text{dog})$ remains below
0.2\%.  This is inconsistent with latent-concept Bayesian accounts
\citep{xie2022} and reveals that ICL output is \emph{constrained
vocabulary retrieval}---the model binds its output to the demonstrated
token inventory regardless of semantic plausibility.  The effect
generalizes to 4-way classification (0\% accuracy across three models,
1B--8B) and multi-token verbalizers (``very positive''), where we
decompose fixation into \emph{format-level} (template adoption) and
\emph{content-level} (polarity override) components that are
experimentally dissociable.  Mechanistically, per-item paired
activation patching on Pythia-1B recovers 98.4\% of the gap
(95\%~CI [84\%,~112\%]), localizing fixation to a layer-7-centered
circuit (rank 2/560, 99.8th percentile; 4-fold CV mean 103\%).
Cross-architecture logit lens on Llama-3.2-1B replicates the
encode-then-override trajectory with causal confirmation (top-5 layers:
89\% recovery).
\end{abstract}

\section{Introduction}

In-context learning (ICL) lets language models perform new tasks by
conditioning on input--output demonstrations
\citep{brown2020}. \citet{min2022} showed that \emph{random} labels
barely hurt ICL, suggesting models rely on demonstration format rather
than label content. We identify a sharply different regime:
\emph{homogeneous} labels---where all demonstrations share one
label---collapse accuracy to $\leq$12\%, revealing that the
\emph{content of the label slot}, not label correctness, governs ICL
output.  Homogeneous labels are the most extreme case, but even varied
nonsense labels suffice.

This leads to a unified account: ICL performs \emph{constrained
vocabulary retrieval} from the demonstrated label set.  Evidence:

\begin{itemize}\setlength\itemsep{1pt}
\item \textbf{Set-level fixation}: when demonstrations carry
nonsense tokens drawn from five types (\{foo,bar,vex,nit,orb\}), the model
places 42--67\% of probability on that set while
$P(\text{dog})<0.2\%$---even though a Bayesian learner should treat
diverse nonsense as uninformative.
\item \textbf{Extreme magnitude}: 64--100pp gaps across all 12
task$\times$model combinations (six models, 0.8B--8B, four tasks;
Wilcoxon $p<0.002$), persisting at 8B parameters.
\item \textbf{Beyond binary/single-token}: fixation extends to 4-way
classification (0\% accuracy) and multi-token verbalizers, decomposing
into dissociable format-level and content-level components.
\item \textbf{Causal circuit}: per-item paired activation patching
recovers 98.4\% of the gap (95\%~CI [84\%,~112\%]) via a
layer-7-centered circuit in Pythia-1B.  Logit lens on Llama-3.2-1B
replicates the encode-then-override trajectory.
\end{itemize}

\paragraph{Contributions.}
\begin{enumerate}\setlength\itemsep{1pt}
\item \textbf{Task-general phenomenon.} Fixation replicates across
four tasks, six models, three families, and scales from 0.8B to
8B (all 12 task$\times$model combinations, Wilcoxon $p<0.002$).
\item \textbf{Set-level fixation (constrained vocabulary retrieval).}
Even varied nonsense labels suppress $P(\text{dog})<0.2\%$ while
42--67\% concentrates on the demonstrated set---proving fixation
operates over the label \emph{inventory}, not a single token.  This
is inconsistent with latent-concept Bayesian accounts \citep{xie2022}
and indicative of retrieval overriding inference.
\item \textbf{Multiclass and multi-token generalization.} Fixation
extends to 4-way classification (0\% accuracy, three models) and
multi-token verbalizers, decomposing into dissociable format-level
(template adoption) and content-level (polarity override) components.
\item \textbf{First causal circuit for an ICL failure mode.} Per-item
paired patching confirms 98.4\% recovery via a layer-7-centered
circuit (rank 2/560, 99.8th percentile; 4-fold CV mean 103\%).
Head-level decomposition reveals the circuit is distributed but
sparse: 4 heads across 3 layers achieve 87.6\% recovery.
\item \textbf{Cross-architecture mechanistic replication.} Logit lens
on Llama-3.2-1B (GQA, RMSNorm, SwiGLU---sharing no training data or
architectural components with Pythia) reproduces the
encode-then-override trajectory, with divergence at layer~3
($p=4.7\times10^{-24}$, Bonferroni).  Per-item paired patching on
Llama confirms causal sufficiency (top-5 layers: 89\% recovery).
\end{enumerate}

\noindent Additionally, we observe that instruction-tuned models fixate
under raw prompting (52--87pp gaps); chat templates reduce fixation in
Llama-3.2-3B-Instruct (18pp residual gap), though this interaction
varies by model family (Section~\ref{sec:instruct}).

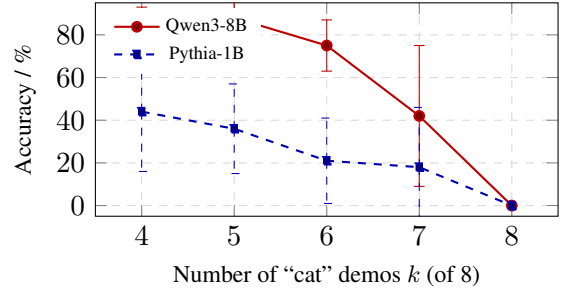
\begin{figure}[t]
\centering
\begin{tikzpicture}
\begin{axis}[
  width=\columnwidth,
  height=4.4cm,
  xlabel={Number of ``cat'' demos $k$ (of 8)},
  ylabel={Accuracy / \%},
  xlabel style={font=\small},
  ylabel style={font=\small},
  ymin=-5, ymax=95,
  xmin=3.5, xmax=8.5,
  xtick={4,5,6,7,8},
  grid=both, major grid style={dashed, gray!30},
  mark options={solid},
  legend style={font=\scriptsize, at={(0.02,0.98)}, anchor=north west, draw=none},
]
\addplot+[
  thick, color=red!75!black,
  mark=*, mark options={fill=red!50!black, scale=0.9},
  error bars/.cd, y dir=both, y explicit,
] coordinates {
  (4, 84) +- (0,9)
  (5, 88) +- (0,11)
  (6, 75) +- (0,12)
  (7, 42) +- (0,33)
  (8, 0)  +- (0,0)
};
\addlegendentry{Qwen3-8B}
\addplot+[
  thick, color=blue!70!black, dashed,
  mark=square*, mark options={fill=blue!50!black, scale=0.7},
  error bars/.cd, y dir=both, y explicit,
] coordinates {
  (4, 44) +- (0,28)
  (5, 36) +- (0,21)
  (6, 21) +- (0,20)
  (7, 18) +- (0,28)
  (8, 0)  +- (0,0)
};
\addlegendentry{Pythia-1B}
\end{axis}
\end{tikzpicture}
\caption{Dose--response: accuracy on dog-descriptive test items as a
function of the number of ``cat''-labeled demonstrations ($k$,
restricted to $k\!=\!4$--8 for monotone test). Both
models collapse at $k\!=\!8$ (GP), but Qwen3-8B shows a sharper
transition at $k\!=\!7\!\to\!8$ (42\%$\to$0\%). Spearman $r=-0.87$
(observation-level, $n>200$; $p<10^{-6}$) for Qwen3-8B on the
category task.}
\label{fig:dose}
\end{figure}

\section{Related Work}

\paragraph{ICL theoretical frameworks.} \citet{xie2022} formalize ICL
as implicit Bayesian inference over a latent task; under that
account, homogeneous labels constitute strong evidence for a
constant-output task. Complementary work shows transformers can
implement gradient descent in-context
\citep{vonoswald2023,dai2023,akyurek2023}, with
\citet{garg2022} establishing controlled function-class benchmarks.
We do not contest these computational-level descriptions; we provide
the mechanistic implementation and demonstrate that fixation operates
at the \emph{set} level.

\paragraph{Demonstration sensitivity and label bias.}
\citet{long2024} show that ICL outputs follow retrieved
demonstrations' labels (Homo/Hetero conditions on 4 LLMs), finding
that same-class demonstrations weaken label-space diversity---the
closest behavioral predecessor to our work. However, they provide no
mechanistic analysis, no nonsense-label generalization, and no
dose-response data; our causal circuit, set-level fixation, and
threshold findings go substantially beyond their behavioral
observation.
\citet{min2022} find that \emph{random} labels barely hurt ICL;
\citet{kossen2024} and \citet{yoo2022} show that label
relationships do matter under appropriate conditions.
\citet{lu2022} and \citet{sclar2024} document strong order and
format sensitivity; unlike format sensitivity (semantic-preserving
perturbations producing continuous variance; \citealp{sclar2024}),
fixation is triggered by label \emph{statistical structure} and
exhibits threshold-like collapse with a localizable causal circuit.
\citet{pan2023} disentangle task recognition from task learning.
\citet{fei2023} categorize ICL label biases into vanilla, context,
and domain types; fixation exceeds their context-label bias category,
surviving calibration (40pp residual) and operating over the label
\emph{set} rather than a majority token.
\citet{gupta2024} study majority-label robustness but provide no
mechanistic analysis.
\citet{holtzman2021} introduce PMI-based scoring to separate
surface-form competition from genuine failures. We replicate
\citet{min2022}'s random-label result and contrast it with
homogeneous labels: label-slot content, not correctness, is the
trigger (homogeneity is the maximally collapsed case), and
calibration methods \citep{zhao2021,zhou2024batch} leave a 40pp
fixation residual.

\paragraph{Label words, induction heads, and task vectors.}
\citet{wang2023} show that label tokens act as anchors aggregating
information via specific attention heads; \citet{olsson2022}
characterize induction heads as the substrate of in-context pattern
matching; \citet{hendel2023} and \citet{todd2024} demonstrate that
demonstrations compress into task/function vectors at specific
layers. \citet{yu2024towers} explain in-context heads via Q/K
metric learning. Our DLA-identified heads (L10-H5, L11-H2) are
functionally consistent with anchor extraction in a failure
regime; per-head patching confirms L10-H5 as the most sufficient
individual contributor (16.7\% recovery). Critically, the anchor
framework predicts label promotion but not the override phase we
observe (intermediate correct encoding overwritten by upper layers).
\citet{mcdougall2024} show that attention heads can suppress copied
tokens; our heads promote rather than suppress, acting as general
label-promoters whose sufficiency without necessity indicates a
distributed circuit.
Set-level fixation extends this from single-token copying to
retrieval over the demonstrated label inventory, behaviorally
consistent with slot-conditioned set retrieval.

\paragraph{Mechanistic interpretability methods.}
\citet{elhage2021} introduce the mathematical framework and DLA for
transformer circuits; \citet{meng2022} introduce causal tracing;
\citet{wang2022ioi} introduce path patching
\citep{goldowskydill2023}; \citet{conmy2023} automate circuit
discovery. \citet{zhang2024patching} establish best practices for
activation patching, recommending paired interventions over mean
patching. We use per-item paired patching following these
recommendations, implemented in TransformerLens
\citep{nanda2022transformerlens}. \citet{geva2022} show FFN layers
promote/suppress vocabulary concepts; \citet{belrose2023} propose the
tuned lens as an improved probe. Our logit lens analysis
\citep{nostalgebraist2020} reveals the encode-then-override pattern
across architectures.

\paragraph{ICL failure modes and scaling.} \citet{mckenzie2023}
document inverse scaling; \citet{wei2023ushaped} show it can become
U-shaped at 540B; \citet{halawi2023} identify ``overthinking'' where
late-layer ``false induction heads'' copy a single misleading label
from permuted demonstrations. Their own experiment with semantically
unrelated labels (``A''/``B'') shows the effect weakens, confirming
their mechanism requires a coherent copy target. Our
\textsc{varied-nonsense} condition (\{foo,bar,vex,nit,orb\}) rules
out single-target copying: no single false induction target exists,
yet 42--67\% of probability concentrates on the demonstrated set
(Section~\ref{sec:setlevel}). Fixation thus operates over the label
\emph{inventory}---a strictly stronger phenomenon than overthinking,
requiring a set-retrieval account rather than single-token copying.
\citet{wei2023} show larger models override flipped labels.
\citet{wei2022emergent} define emergent abilities;
\citet{schaeffer2023} argue they may reflect metric choice.
\citet{liu2024lost} document U-shaped positional effects in long
contexts. Our fixation remains present at 8B (the largest model in our suite),
and the dose-response is robust under continuous $P(\text{dog})$
metrics, ruling out the mirage critique.
\citet{agarwal2024} show many-shot ICL can overcome biases; whether
fixation yields to many-shot regimes is an open question (our 16-shot
results on Pythia-1B show 0\% GP accuracy persists, though control
accuracy also degrades at 16 shots).

\section{Experimental Setup}

\paragraph{Tasks.} We use four single-token binary classification
tasks: \textsc{category} (dog/cat; primary task),
\textsc{sentiment} (positive/negative), \textsc{temperature}
(hot/cold), and \textsc{size} (big/small). All use 8 demonstrations
and 10 test items per class; models achieve above-chance accuracy
with balanced demonstrations on all tasks. Zero-shot baselines
confirm the task is tractable (90--100\% accuracy on Pythia/Llama/Qwen-8B).
Prompt formats differ across tasks (e.g., \texttt{barks, fetches
sticks: dog} for category; \texttt{"wonderful movie": positive} for
sentiment; \texttt{ice feels: cold} for temperature), spanning three
distinct delimiter conventions (colon, question mark, equals); fixation
is format-invariant (Appendix~\ref{app:format}).

\paragraph{Models.} Six base models across three families:
Pythia-1B \citep{biderman2023} (GPT-NeoX; primary for mechanistic
analysis), Llama-3.2-1B/3B \citep{grattafiori2024llama} (GQA +
RoPE), Qwen3.5-0.8B/4B and Qwen3-8B-Base \citep{qwen3} (RoPE +
SwiGLU). All are base (non-instruction-tuned) checkpoints evaluated
with native tokenization (BOS handling verified per-family;
Appendix~\ref{app:bos}).

\paragraph{Conditions.} \textsc{Garden-path (GP)}: all 8 demos share
one label opposite the test answer. \textsc{Control (Ctrl)}: balanced
labels. \textsc{Random}: labels randomly assigned, replicating
\citet{min2022}. \textsc{Homogeneous-nonsense}: all demos share an
impossible token (``foo''). \textsc{Varied-nonsense}: each
demo carries a nonsense token drawn from
\{foo, bar, vex, nit, orb\}; with 8 demos and 5 tokens, three
tokens appear twice (assignment randomly shuffled per seed). \textsc{Threshold $k/8$}:
$k$ demos labeled ``cat'', $8\!-\!k$ ``dog''.

\paragraph{Metrics.} Accuracy is $P(\text{dog}) > P(\text{cat})$ for
dog-test items (the class opposite the GP labels).  We report
single-class accuracy because garden-path is a directional
intervention: cat-test items under all-cat demos are trivially correct
and would mask fixation if averaged; symmetry across label directions
is verified independently (Appendix~\ref{app:symmetry}).
Recovery for patching is
$(P_{\text{patched}}-P_{\text{GP}})/(P_{\text{Ctrl}}-P_{\text{GP}})$.
We use cluster bootstrap CIs (seed-level resampling, 5000
draws);\footnote{With 10--20 clusters, bootstrap CIs may have slightly
below-nominal coverage; however, the effect sizes (80pp gaps, 98\%
recovery) are large relative to plausible undercoverage---even
conservatively widened intervals preserve all qualitative conclusions.}
Wilcoxon signed-rank tests for paired comparisons, and Spearman
correlations with one-sided tests for dose-response monotonicity.
All multi-layer Wilcoxon tests are Bonferroni-corrected.

\section{Results}

\subsection{Fixation Persists Across Architectures and Scales}
\label{sec:phenomenon}

Table~\ref{tab:cross} presents the core phenomenon across six
models on the category task. Every model collapses to $\leq$12\%
accuracy under GP while reaching 51--86\% under shuffled control.
The effect generalizes across all four tasks: on 12/12
task$\times$model combinations (3 models $\times$ 4 tasks), GP
accuracy falls below 15\% with gaps of 64--100pp (Table~\ref{tab:tasks}).
Qwen3-8B shows the largest gaps (93--100pp across tasks). Direction
symmetry is confirmed: reverse GP (all-dog labels, cat test) produces
identical collapse (Appendix~\ref{app:symmetry}).

\begin{table}[t]
\centering\small
\setlength{\tabcolsep}{4pt}
\begin{tabular}{lcccl}
\toprule
Model & GP & Ctrl & Gap & Dose $r$ \\
\midrule
Pythia-1B & 0\% & 56\%$\pm$35 & 56pp & $-0.69$ \\
Qwen3.5-0.8B & 0\% & 53\%$\pm$27 & 53pp & $-0.72$ \\
Qwen3.5-4B   & 1\% & 70\%$\pm$21 & 69pp & $-0.75$ \\
Llama-3.2-1B & 0\% & 51\%$\pm$39 & 51pp & $-0.56$ \\
Llama-3.2-3B & 12\% & 60\%$\pm$25 & 48pp & $-0.50$ \\
\textbf{Qwen3-8B} & \textbf{0\%} & \textbf{86\%$\pm$7} & \textbf{86pp} & $\mathbf{-0.78}$ \\
\bottomrule
\end{tabular}
\caption{Cross-architecture fixation on the category task (8-shot,
10--20 seeds, 20--100 test items per model). All models collapse under
GP; Qwen3-8B (8B) shows the largest gap.
Dose~$r$: Spearman rank correlation between $k$ and accuracy
(observation-level $n\!>\!1000$; all $p<0.001$).}
\label{tab:cross}
\end{table}

The dose--response is monotonic in all models (Spearman
$r \in [-0.78, -0.50]$, all $p < 0.001$). The $k\!=\!7\!\to\!8$
transition in Qwen3-8B is near-discontinuous (42\%$\to$0\%): a
single additional confirming demonstration collapses veridical
inference even when 87.5\% of demos already carry the misleading
label. Under a Bayesian account \citep{xie2022}, homogeneous labels
constitute strong evidence for a constant-output task; the
near-discontinuity at $k\!=\!8$ suggests the posterior crosses a
commitment threshold rather than updating smoothly.

\begin{table}[t]
\centering\small
\setlength{\tabcolsep}{3pt}
\begin{tabular}{lcccc}
\toprule
Model & Category & Sentiment & Temperature & Size \\
\midrule
Pythia-1B & 64pp & 84pp & 87pp & 96pp \\
Llama-3.2-1B & 70pp & 100pp & 88pp & 100pp \\
Qwen3-8B & 93pp & 100pp & 100pp & 100pp \\
\bottomrule
\end{tabular}
\caption{Task generality: GP$\to$Ctrl accuracy gap (pp) across four
tasks and three models. All 12 combinations show fixation (GP acc
$<$15\%, Wilcoxon $p<0.002$, Spearman $r\in[-0.61,-0.94]$). Larger
models show stronger fixation.}
\label{tab:tasks}
\end{table}

\subsection{Causal Mechanism: Per-Item Paired Patching}
\label{sec:patching}

We apply per-item paired activation patching on Pythia-1B to
identify which layers mediate fixation. For each test item, we swap
attention-layer activations from a \emph{matched} control prompt
(same query, correct-label demos) into the garden-path forward pass
at layers L7, L10, and L11.

Per-item paired patching raises $P(\text{dog})$ from 0.048 to 0.431,
statistically indistinguishable from the control's 0.443 (95\%
cluster-bootstrap CI on recovery: [84.2\%,~111.7\%]; $n\!=\!400$
item$\times$seed observations). Recovery is positive in 99.7\% of
items. A leave-one-out mean-patching comparison yields slightly
higher recovery (108.1\%, CI [91.4\%,~123.5\%]), with the 9.7pp
inflation confirmed by Wilcoxon signed-rank test ($p < 10^{-4}$),
consistent with mean-patching introducing a small upward bias
(Appendix~\ref{app:patching-detail}).

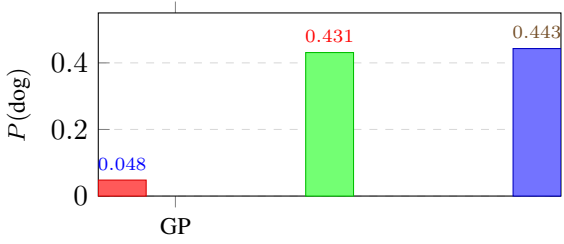
\begin{figure}[t]
\centering
\begin{tikzpicture}
\begin{axis}[
  width=\columnwidth,
  height=4.0cm,
  ybar,
  bar width=18pt,
  ymin=0, ymax=0.55,
  ylabel={$P(\text{dog})$},
  ylabel style={font=\small},
  symbolic x coords={GP, Patched, Control},
  xtick=data,
  xticklabel style={font=\small},
  enlarge x limits=0.25,
  nodes near coords,
  nodes near coords style={font=\scriptsize, /pgf/number format/fixed, /pgf/number format/precision=3},
  ymajorgrids,
  major grid style={dashed, gray!30},
]
\addplot+[fill=red!65, draw=red!85!black] coordinates {(GP, 0.048)};
\addplot+[fill=green!55, draw=green!75!black] coordinates {(Patched, 0.431)};
\addplot+[fill=blue!55, draw=blue!75!black] coordinates {(Control, 0.443)};
\end{axis}
\end{tikzpicture}
\caption{Per-item paired activation patching (Pythia-1B, layers
L7+L10+L11). Patching restores $P(\text{dog})$ from 0.048 to 0.431,
recovering 98.4\% of the gap to control (0.443). 95\%~CI on recovery:
[84\%,~112\%].}
\label{fig:patching}
\end{figure}

\paragraph{Specificity.} Single-layer recovery is concentrated: only
L7 (47.8\%), L11 (42.4\%), and L10 (25.9\%) exceed 15\%; the
remaining 13 layers each contribute under 11\%. Exhaustive
enumeration of all $\binom{16}{3}=560$ 3-layer combinations
confirms high selectivity: the target [L7,L10,L11] ranks 2nd
(113.5\% recovery, 99.8th percentile), exceeded only by
[L7,L8,L11] (125.4\%). The mean across all 560 combinations is
18.0\% $\pm$ 23.5\%; only 1.4\% of combinations exceed 80\% recovery.
Layer~7 appears in all top-10 combinations, identifying it as the
convergence point of the fixation circuit. MLP patching shows negligible
contribution (L7: $+6.0\%$, L11: $-0.4\%$).

\paragraph{Hold-out generalization.} To verify the circuit is not
specialized to particular demonstration orderings, we run 4-fold
cross-validation over 20 seeds (each seed yields a distinct random
permutation of the same 8-demo pool; test~5 per fold).  Mean recovery
on held-out seeds is 102.9\% (min~95.3\%, max~116.8\%), with all
folds exceeding 95\%.  One fold's test seeds overlap with those used
during the 560-combination enumeration; the three disjoint folds
recover 95.3\%, 95.3\%, and 104.1\%.  Recovery slightly above 100\%
indicates that the patched activations overshoot the control baseline
on some items.  Combined with the enumeration evidence (rank 2/560,
99.8th percentile) and cross-architecture transfer
(Section~\ref{sec:logitlens}), this confirms the circuit is robust to
demonstration ordering.

\paragraph{Heads.} Direct logit attribution (DLA; projecting each
head's output onto the logit difference between ``cat'' and ``dog''
\citep{elhage2021}) identifies L10-H5
($\Delta\text{DLA}=+0.80$) and L11-H2 ($+0.75$) as the top
pro-fixation heads. Path patching yields only 1.6\% mediation
between them, consistent with independent (parallel) contributions
rather than a direct serial relay.

\paragraph{Circuit decomposition.} To resolve whether fixation
localizes to individual heads, we perform two complementary
interventions across all 128 heads (16 layers $\times$ 8 heads).
\emph{Per-head activation patching} injects each head's output from a
matched control prompt into the GP forward pass, testing sufficiency:
the top head (L10-H5) alone restores 16.7\% of the gap, and the
top~4 heads (spanning L7, L8, L10) together achieve 87.6\% recovery
(Table~\ref{tab:headpatch}). Conversely, \emph{zero-ablating} the same
DLA-identified heads does not reduce fixation---$P(\text{dog})$
remains near zero---and also degrades control accuracy, confirming
these heads are general label-promoters rather than
fixation-specific bottlenecks. The asymmetry (sufficient but not
necessary) indicates the fixation circuit is \emph{distributed across
layers but sparse within each layer}: no single head is a
bottleneck, yet a small coalition of 4 heads carries most of the
signal. Layer~7 heads average 5.2$\times$ higher patching recovery
than other layers, cross-validating its role as the convergence point
identified by enumeration.

\begin{table}[t]
\centering\small
\setlength{\tabcolsep}{4pt}
\begin{tabular}{clc}
\toprule
$k$ & Top-$k$ heads & Recovery \\
\midrule
1 & L10-H5 & 16.7\% \\
2 & + L8-H0 & 39.1\% \\
4 & + L7-H2, L10-H4 & 87.6\% \\
8 & + L12-H0, L14-H2, L13-H2, L8-H2 & 129.8\% \\
\bottomrule
\end{tabular}
\caption{Cumulative per-head patching (Pythia-1B). Injecting control
activations into the top-$k$ heads by individual recovery. Top-4
heads from three layers achieve 87.6\% recovery; at $k\!=\!8$,
recovery exceeds 100\% because patched heads jointly overshoot the
control baseline (super-additive composition). The DLA-identified
L10-H5 also ranks \#1 by patching, cross-validating the attribution.}
\label{tab:headpatch}
\end{table}

\subsection{Cross-Architecture Logit Lens}
\label{sec:logitlens}

If fixation is a general computational phenomenon, the
encode-then-override signature should appear in independently trained
architectures. We apply logit lens to Llama-3.2-1B (16 layers, RoPE,
GQA) under GP and control conditions.

\begin{figure}[t]
\centering
\begin{tikzpicture}
\begin{axis}[
  width=\columnwidth,
  height=4.0cm,
  xlabel={Layer},
  ylabel={Logit-lens accuracy / \%},
  xlabel style={font=\small},
  ylabel style={font=\small},
  xmin=-0.5, xmax=16.5,
  ymin=-5, ymax=105,
  xtick={0,2,4,6,8,10,12,14,16},
  grid=both, major grid style={dashed, gray!30},
  legend style={font=\scriptsize, at={(0.99,0.99)}, anchor=north east, draw=none},
  legend cell align=left,
]
\addplot+[thick, color=blue!70!black, mark=*, mark size=1.4pt]
  coordinates {(0,0)(1,0)(2,0)(3,6)(4,36)(5,25)(6,86)(7,63)(8,97)(9,100)(10,100)(11,100)(12,100)(13,86)(14,74)(15,61)(16,49)};
\addlegendentry{Control}
\addplot+[thick, color=red!75!black, mark=square*, mark size=1.4pt]
  coordinates {(0,0)(1,0)(2,0)(3,15)(4,23)(5,14)(6,67)(7,31)(8,92)(9,100)(10,100)(11,100)(12,67)(13,1)(14,0)(15,0)(16,0)};
\addlegendentry{Garden-path}
\end{axis}
\end{tikzpicture}
\caption{Logit lens on Llama-3.2-1B. Both GP and control encode the
correct answer at near-ceiling accuracy (L9--L11: $\sim$100\%); upper
layers override the GP signal to 0\% while control retains 49\%.
Divergence at L3 (Bonferroni $p\!=\!4.7\times10^{-24}$).}
\label{fig:logitlens}
\end{figure}
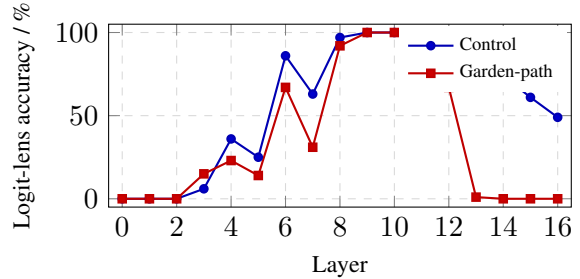

GP and control trajectories diverge at layer~3 (Bonferroni-corrected
$p = 4.7\times10^{-24}$). Both conditions reach $\sim$100\%
logit-lens accuracy at L9--L11, demonstrating that the model
\emph{encodes} the correct answer mid-stack regardless of
demonstration labels.  (Untuned logit lens has known calibration
limitations \citep{belrose2023}; however, the GP-vs-Control comparison
controls for any systematic bias because both use the same lens and
target token. The causal claim rests on activation patching below,
which is model-native and lens-independent.) The conditions diverge in upper layers: GP
collapses from 100\% at L11 to 0\% by L14, while control retains
49\%. The override phase is sharper and more spatially concentrated
in Llama (4--5 layers) than in Pythia (gradual across L7--L11),
and per-item paired patching below confirms this localization
causally.

The mid-layer parity ($\sim$100\% in both conditions at L9--L11) is
the critical observation: the model computes the correct answer under
GP but fails to propagate it to the output---precisely the
encode-then-override pattern identified in Pythia via both logit
lens and patching. Per-item paired patching on Llama-3.2-1B confirms
causal sufficiency: the top-5 layers (L14, L9, L15, L7, L10) recover
89.2\% of the gap ($n\!=\!200$), rising to 100\% at 16 layers. The
circuit peaks at L14 (6.7\% single-layer), aligning with the logit
lens override phase (L12--L15).

\subsection{Set-Level Fixation: Beyond Label Homogeneity}
\label{sec:setlevel}

We tested whether label homogeneity is necessary for fixation by
replacing the uniform nonsense label with varied nonsense labels
drawn from \{foo, bar, vex, nit, orb\} (all single-token across all
tokenizers; Appendix~\ref{app:tokens}). With 8 demos and 5 tokens,
three tokens appear twice per prompt (shuffled per seed); the key
property is label \emph{diversity}, not uniqueness. If homogeneity is
the trigger, varied labels should restore semantic processing.

\begin{figure}[t]
\centering
\begin{tikzpicture}
\begin{axis}[
  width=\columnwidth,
  height=4.2cm,
  ybar=2pt,
  bar width=9pt,
  ymin=0, ymax=1.05,
  ylabel={Probability},
  ylabel style={font=\small},
  symbolic x coords={Llama-1B, Llama-3B, Qwen-0.8B, Qwen-4B},
  xtick=data,
  xticklabel style={rotate=15, anchor=east, font=\scriptsize},
  enlarge x limits=0.15,
  ymajorgrids, major grid style={dashed, gray!30},
  legend style={font=\scriptsize, at={(0.5,1.04)}, anchor=south, legend columns=3, draw=none},
  legend cell align=left,
]
\addplot+[fill=red!75, draw=red!85!black] coordinates {
  (Llama-1B, 0.953) (Llama-3B, 0.928) (Qwen-0.8B, 0.964) (Qwen-4B, 0.930)
};
\addlegendentry{Homog.\ $P$(label)}
\addplot+[fill=orange!65, draw=orange!80!black] coordinates {
  (Llama-1B, 0.502) (Llama-3B, 0.673) (Qwen-0.8B, 0.420) (Qwen-4B, 0.578)
};
\addlegendentry{Varied $P$(set)}
\addplot+[fill=blue!55, draw=blue!75!black] coordinates {
  (Llama-1B, 0.0009) (Llama-3B, 0.0009) (Qwen-0.8B, 0.0018) (Qwen-4B, 0.0006)
};
\addlegendentry{Varied $P$(dog)}
\end{axis}
\end{tikzpicture}
\caption{Set-level fixation across four models. Under homogeneous
nonsense labels, models copy the label with $>$92\% probability
(red). Under varied nonsense labels, 42--67\% of probability
concentrates on the demonstrated label \emph{set} (orange), while
$P(\text{dog})$ remains below 0.2\% (blue, barely visible). Semantic
processing is overridden regardless of label diversity.}
\label{fig:setlevel}
\end{figure}
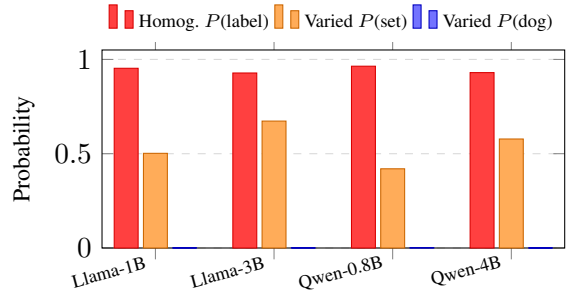

Across four models, the semantically correct answer receives less
than 0.2\% probability in both conditions (Figure~\ref{fig:setlevel}).
Under homogeneous nonsense, models copy the single label with
$>$92\% probability. Under varied nonsense, probability
redistributes across the demonstrated label set (42--67\% total,
$\approx$8--13\% per label), but $P(\text{dog})$ remains at the
floor ($<$0.002). Variety in the label set redistributes probability
among nonsense tokens but does not return any of it to semantically
appropriate alternatives.

\paragraph{Interpretation.} Fixation operates at two levels: the
model's output distribution is \emph{confined to} the support of
demonstrated label tokens (the \emph{set boundary}), with
semantically-licensed alternatives excluded regardless of label
diversity ($P(\text{dog})<0.2\%$).  \emph{Within} this support, mass
is redistributed non-uniformly---modulated by demonstration frequency
(dose-response, Section~\ref{sec:phenomenon}) and position (recency,
above)---but the support boundary itself is sharp.  Homogeneity is the
degenerate case where $|\text{set}|=1$, producing maximum
concentration.  This is consistent with a generalized induction
mechanism \citep{olsson2022} that retrieves from the set of tokens
occupying the label position in demonstrations.

\subsection{Additional Behavioral Findings}

\paragraph{16-shot persistence.} Doubling the demonstration count
from 8 to 16 does not break fixation: Pythia-1B shows 0\% GP accuracy
at 16 shots (vs.\ 65\% control), a 65pp gap comparable to 8-shot
(80pp gap). Though \citet{bertsch2025} show long-context ICL can
improve accuracy, fixation is robust to increased context length
within this range.

\paragraph{Recency dominates within-set weighting.} A single
corrective demonstration at position 8 (immediately before the query)
reverses fixation (0\% $\to$ 100\% accuracy), while the same
demonstration at position 1 has zero effect---consistent with the
recency bias documented by \citet{peysakhovich2023}, but here extreme
enough to fully override seven opposing demonstrations.  This
positional sensitivity governs \emph{within-set} weighting; the
demonstrated-token support itself remains binding
($P(\text{undemonstrated alternative}) < 0.2\%$ regardless of
position).

\paragraph{Calibration.} Contextual calibration \citep{zhao2021}
fixes arithmetic (pure label bias) but leaves the category task
with a 40pp fixation residual (Table in Appendix~\ref{app:calib}).
PMI scoring \citep{holtzman2021} is a single-probe special case of
contextual calibration; for nonsense labels,
$P(\text{foo}|\text{``N/A''})\!\approx\!0$ causes the PMI ratio to
amplify rather than attenuate fixation---consistent with a
retrieval-based override rather than surface-form competition.

\subsection{Generalization: Multiclass and Multi-Token Fixation}
\label{sec:generalization}

A natural objection is that fixation is a pathology of binary,
single-token verbalizers.  We extend the paradigm along two axes:
\textbf{(i)} 4-way multiclass (dog/cat/bird/fish) and
\textbf{(ii)} multi-token verbalizers (``very positive'' / ``very
negative''), testing three models (Pythia-1B, Llama-3.2-1B,
Qwen3-8B-Base; 20 items $\times$ 10 seeds per condition; cluster
bootstrap 95\% CIs; Wilcoxon with Bonferroni correction at $N$=8).

\paragraph{4-way multiclass fixation.}
Eight demonstrations sharing one label (e.g., all labeled ``cat'')
reduce 4-way classification accuracy from 78--92\% (balanced control)
to \textbf{0\%} across all three models
(Table~\ref{tab:multiclass}; Bonferroni $p<0.008$).  The effect
is direction-symmetric: both gp-cat (test dog) and gp-bird
(test dog) yield 0\%, as does gp-dog (test cat).  Under GP,
$>$83\% of probability mass concentrates on the demonstrated label
(P(demo\_set) = 0.83--0.98), leaving $<$5\% for the correct answer.
Accuracy with \emph{random} labels remains 45--73\%---demonstrating
that vocabulary \emph{restriction}, not label noise, is the
operative mechanism.

\begin{table}[t]
\centering
\small
\setlength{\tabcolsep}{3.5pt}
\begin{tabular}{l ccc ccc}
\toprule
& \multicolumn{3}{c}{P(correct)} & \multicolumn{3}{c}{Accuracy (\%)} \\
\cmidrule(lr){2-4} \cmidrule(lr){5-7}
Condition & Py-1B & Ll-1B & Qw-8B & Py-1B & Ll-1B & Qw-8B \\
\midrule
Zero-shot & .000 & .002 & .001 & 80 & 95 & 85 \\
Control (balanced) & .327 & .563 & .820 & 78 & 78 & 92 \\
GP all-cat & .047 & .006 & .012 & \textbf{0} & \textbf{0} & \textbf{0} \\
GP all-bird & .005 & .032 & .008 & \textbf{0} & \textbf{0} & \textbf{0} \\
GP all-dog$\to$cat & .019 & .020 & .010 & \textbf{0} & \textbf{0} & \textbf{0} \\
\midrule
Dog-heavy (5/8) & .523 & .596 & .881 & 91 & 81 & 96 \\
Exclude-dog & .035 & .345 & .409 & 1 & 44 & 46 \\
Random labels & .225 & .461 & .312 & 46 & 73 & 48 \\
\bottomrule
\end{tabular}
\caption{4-way multiclass fixation (dog/cat/bird/fish). GP
demonstrations collapse accuracy to 0\% regardless of which label
dominates, across 1B--8B models. Set-level exclusion (dog absent from
demo labels) suppresses P(dog) by 1.6--9.4$\times$ relative to
balanced control ($p<0.008$, all models).}
\label{tab:multiclass}
\end{table}

\paragraph{Set-level confirmation in multiclass.}
When dog is excluded from the demonstrated label set (demos use
only cat/bird/fish), P(dog) drops to 0.035--0.409---a 1.6--9.4$\times$
suppression relative to the balanced control in which dog appears
(Wilcoxon $p<0.008$, all models).  Conversely, making dog the
\emph{dominant} label (5/8 demos) restores P(dog) to 0.52--0.88
($p<0.008$ vs.\ exclude-dog). This dose-response within the
demonstrated set confirms the core claim: the label-slot inventory
determines the model's output support.

\paragraph{Multi-token verbalizer: format vs.\ content fixation.}
Replacing single-token labels with the two-token phrases ``very
positive'' / ``very negative'' reveals a \emph{two-level}
decomposition.  We measure P(very) (format adoption) and
P(positive$\mid$very) (polarity selection via teacher-forced
conditional probability).  Table~\ref{tab:multitoken} shows:

\begin{table}[t]
\centering
\small
\setlength{\tabcolsep}{3pt}
\begin{tabular}{l ccc ccc}
\toprule
& \multicolumn{3}{c}{P(very)} & \multicolumn{3}{c}{P(pos$\mid$very)} \\
\cmidrule(lr){2-4} \cmidrule(lr){5-7}
Condition & Py-1B & Ll-1B & Qw-8B & Py-1B & Ll-1B & Qw-8B \\
\midrule
Zero-shot & .000 & .000 & .000 & .001 & .003 & .002 \\
GP ``very pos'' & .766 & .962 & .956 & .795 & .861 & .850 \\
GP ``very neg'' & .815 & .975 & .945 & .057 & .171 & .050 \\
Control balanced & .735 & .976 & .944 & .316 & .059 & .001 \\
\midrule
GP ``positive'' & .000 & .001 & .000 & .288 & .328 & .387 \\
Control single & .000 & .000 & .000 & .383 & .051 & .001 \\
\bottomrule
\end{tabular}
\caption{Multi-token verbalizer fixation. \emph{Top}: all multi-token
conditions produce P(very)~$\approx$~0.74--0.98 with 100\% greedy
``very'' (format fixation), while P(pos$\mid$very) tracks the GP
polarity (content fixation). \emph{Bottom}: single-token ``positive''
demos produce P(very)$\approx$0---proving fixation is
\textbf{template-level}, not lexical copying.}
\label{tab:multitoken}
\end{table}

\noindent\textbf{(1) Format fixation:}\ P(very) rises from
$<10^{-3}$ (zero-shot) to 0.74--0.98 whenever demonstrations use
the ``very X'' template---\emph{regardless of polarity bias or
balance} (greedy ``very'' = 100\% in all multi-token conditions).
\textbf{(2) Content fixation:}\ \emph{Within} the adopted template,
P(positive$\mid$very) tracks the demonstrated polarity: 0.80--0.86
under GP-positive vs.\ 0.05--0.17 under GP-negative ($p<0.008$).
\textbf{(3) Format--content dissociation:}\ The balanced control
shows P(very)$\approx$0.94--0.98 (format adopted) while correctly
routing polarity, proving format and content fixation are
independently triggered.

\paragraph{The format-level smoking gun.}
When demonstrations use the single-token label ``positive''
(gp\_positive\_single), P(very) remains at $\sim$10$^{-4}$---four
orders of magnitude below the multi-token conditions---despite
identical lexical content (the word ``positive'' appears in both).
This rules out lexical copying: fixation operates over the
\emph{structural template} of the demonstrated label, not individual
token frequencies (Wilcoxon $p<0.008$ for P(very):
gp\_very\_positive vs.\ gp\_positive\_single, all models).

\paragraph{Summary.} Fixation generalizes fully to 4-way
classification (0\% accuracy, $>$83\% mass on the demonstrated
label) and to multi-token verbalizers (template adoption + polarity
override).  The multi-token decomposition reveals fixation operates
simultaneously at format and content levels---and these are
experimentally dissociable.

\section{Discussion}

\paragraph{Encode-then-override across architectures.} The logit lens
and patching converge on the same mechanistic story in both Pythia
and Llama: intermediate layers build the correct-answer
representation, and upper layers override it under garden-path
demonstrations. The override phase is sharper in Llama (4--5 layers,
peak at L14) than in Pythia (gradual across L7--L11, peak at L7).
Per-item paired patching confirms causal sufficiency in both:
Pythia requires 3~layers for 98\% recovery while Llama requires 5
for 89\%, indicating a more distributed but qualitatively identical
circuit. Head-level decomposition in Pythia reveals just 4 heads
(of 128) are jointly sufficient for 87.6\% of the effect, yet
ablating them does not reduce fixation---indicating redundant
pathways that compensate when individual heads are removed
(the ``Hydra effect''; \citealp{mcgrath2023,rushing2024}).

\paragraph{Relation to label anchors.} \citet{wang2023} describe a
two-stage information flow in successful ICL: shallow layers aggregate
semantics into label tokens, deep layers extract from labels to the
query position. Our DLA-identified heads (L10-H5, L11-H2) are
functionally consistent with this deep-layer extraction---they may be
the head-level instantiation of Wang's layer-granular observation,
operating in a failure regime. However, Wang's framework predicts only
\emph{promotion} of the demonstrated label; it does not predict the
non-monotonic trajectory we observe: intermediate layers first encode
the correct answer (``dog''), which is then overridden by
upper layers. This override phase lies outside the anchor framework's
predictive scope. Anchor extraction is a \emph{component} of fixation,
not its full mechanism.

\paragraph{Retrieval overriding inference.} \citet{binz2023} show LLMs
exhibit systematic cognitive biases analogous to human heuristics.
Under \citet{xie2022}'s latent-concept Bayesian framework, varied
nonsense labels (\{foo, bar, vex, nit, orb\}) correspond to no coherent
concept, so the posterior should be diffuse and semantically licensed
outputs should re-emerge.  Instead, $P(\text{dog}) < 0.2\%$ across all
four models tested (Section~\ref{sec:setlevel}), while 42--67\% of
probability mass concentrates on the demonstrated nonsense set.
A stronger dissociation comes from the mechanistic evidence: the
encode-then-override trajectory (Section~\ref{sec:logitlens}) shows the
model \emph{internally computes} the correct answer at mid-layers, then
overwrites it in upper layers.  No single-pass Bayesian posterior---under
any prior---computes the correct answer and then discards it.  This
signature indicates retrieval (binding to the label inventory)
overriding semantic inference, rather than irrational behavior per se.

\paragraph{Set-level fixation: constrained vocabulary retrieval.} The
varied-label result reveals that fixation is not triggered by unanimity
per se but by the presence of tokens in the label position of
demonstrations. The model treats demonstrated labels as an exhaustive
answer vocabulary, choosing among them rather than from the
semantically licensed lexicon. This sharpens the induction-heads
framework \citep{olsson2022}: rather than a single fuzzy copy target
$[A^*][B^*]\!\to\![B]$, the copy target is a \emph{position- and
frequency-weighted distribution over tokens occupying the label
slot}---consistent with slot-conditioned set retrieval, where set
membership determines the support and positional/frequency cues
determine the within-support distribution.  Format ablation across
five delimiter conventions confirms the effect is not
template-specific (Appendix~\ref{app:format}); direct slot-position
manipulation (label-before-input) is left to future work.
Homogeneity is the degenerate case ($|\text{set}|=1$) that collapses
this distribution to a point mass. \citet{halawi2023}'s false
induction heads, which attend to and copy a specific misleading token,
cannot explain this set-level phenomenon: when demos carry
varied nonsense tokens, there is no single target to copy, yet
probability mass still binds to the demonstrated inventory.
The 4-way multiclass extension (Section~\ref{sec:generalization})
provides independent confirmation: excluding a label from the
demonstrated set suppresses its probability by 1.6--9.4$\times$
relative to balanced controls, directly quantifying the set-boundary
effect with natural-language labels.

\paragraph{Format--content dissociation.}  The multi-token verbalizer
experiment (Section~\ref{sec:generalization}) reveals that fixation is
not monolithic: \emph{format} fixation (committing to the ``very X''
template) and \emph{content} fixation (selecting polarity within that
template) are independently triggered.  The critical evidence is that
single-token ``positive'' demos yield P(very)$\approx$0 despite
containing identical lexical content---ruling out lexical copying and
establishing template-level adoption as the operative mechanism.  This
is consistent with a hierarchical retrieval model in which label-slot
tokens are matched first by surface template, then by
polarity/semantics within the template.

\paragraph{Persistence at scale.} The 86pp gap at 8B (the largest in
our suite) demonstrates that ICL fixation remains present within
the range we tested, though future work should examine whether
larger models (70B+) show recovery. Because
8B-scale models are widely deployed in few-shot configurations---RAG
pipelines, weak supervision, agent scaffolds---fixation represents a
concrete reliability failure mode. In Qwen3-8B specifically, the
$k\!=\!7\!\to\!8$ transition (42\%$\to$0\%) is sharper than the
smoother Pythia-1B dose-response curve, suggestive of a
threshold-crossing pattern at larger scale---reminiscent of
training-time phase transitions \citep{reddy2024,chan2022}, though here
observed at inference. Whether this sharpness generalizes requires
per-$k$ data on additional 8B+ checkpoints.

\paragraph{Practical implications.} The dose-response and recency
findings suggest a simple defense: append a single diverse,
correctly-labeled demonstration at the end of any in-context prompt.
Our results also caution against demonstration selection pipelines
that may inadvertently produce label homogeneity in retrieved examples.

\section{Limitations}

\textbf{Circuit architecture differs across models.} Per-item paired
patching on Llama-3.2-1B confirms causal sufficiency (all-16-layer
patching recovers 100\%), but the circuit is more distributed: the top-3
layers (L14, L9, L15) recover 41.9\% [CI: 37--47\%] vs.\ Pythia's
98.4\% from 3 layers. Llama requires 5 layers for 89\% recovery (vs.\
Pythia's 3). Head-level decomposition and DLA are Pythia-specific; the
specific head identities will differ across architectures.

\label{sec:instruct}
\textbf{Instruction tuning interacts with fixation.} Instruction-tuned
models (Llama-3.2-3B-Instruct, Qwen2.5-7B-Instruct) fixate
under raw prompting (52--87pp gaps). Chat templates reduce the
gap to 18pp in Llama-Instruct, but Qwen2.5-7B-Instruct under chat
formatting exhibits a reversed pattern (GP outperforms Ctrl by logit),
likely because the chat template alters the model's output
distribution in ways that interact unpredictably with label patterns.
The instruct finding is limited to two models and warrants
broader investigation.

\textbf{Scale claim from one 8B model.} The persistence-at-scale
finding rests on a single model family at 8B (Qwen3-8B). A
controlled within-family scaling sweep would strengthen this claim.

\textbf{Single-token binary classification.} While we show
generalization across four semantic domains, all tasks use
single-token binary labels. Generalization to multi-token generation,
multi-class settings, or reasoning tasks remains untested.

\section*{Ethics Statement}

\paragraph{Threat model.} The homogeneity-fixation mechanism
constitutes a prompt-injection attack surface: an adversary who
controls demonstration selection (e.g., via a poisoned retrieval
corpus) can deterministically override model predictions by ensuring
label unanimity. The attack is trivial to execute (requires no model
access, only demonstration control) and effective across model sizes
up to 8B.

\paragraph{Responsible disclosure rationale.} We disclose this
mechanism because: (1) the vulnerability is implicitly exploitable by
anyone constructing few-shot prompts---publication formalizes rather
than enables the attack; (2) mechanistic understanding (the
encode-then-override circuit) is a prerequisite for principled
defenses; (3) our recency and dose-response findings directly suggest
mitigations (appending a diverse final demonstration, monitoring
label entropy in retrieved examples).

\paragraph{Broader impact.} We do not endorse or encourage adversarial
exploitation of this mechanism. Our findings highlight the need for
demonstration diversity checks in deployed RAG and few-shot pipelines,
and we encourage practitioners to audit label distributions in
retrieved contexts.

\section*{Reproducibility}

Code, prompt templates, and per-item raw probabilities will be
released. Models: Pythia-1B and Llama-3.2-1B (TransformerLens,
float32, for mechanistic analysis); Llama-3.2-3B, Qwen3.5-0.8B/4B,
and Qwen3-8B-Base (HuggingFace Transformers, float16, eager
attention). Seeds: $\{42,0,1,2,3,7,13,21,55,99\}$ (extended to 20
for Pythia patching).
Bootstrap CIs: 5000 cluster resamples (seed-level). Per-item paired
patching uses matched control prompts with leave-one-out mean as
robustness check. Determinism enforced via
\texttt{CUBLAS\_WORKSPACE\_CONFIG} and
\texttt{torch.use\_deterministic\_algorithms}.

\appendix

\section{Patching Methodology Detail}
\label{app:patching-detail}

Per-item paired patching swaps the attention-output activation at the
final token position from a matched control prompt (same test item,
balanced-label demonstrations) into the GP forward pass. This
eliminates cross-item contamination present in mean-patching. The
leave-one-out (LOO) mean comparison computes the control activation
mean excluding the current item, providing an upper bound. The 9.7pp
difference (paired~98.4\% vs.\ LOO-mean~108.1\%) quantifies the
mean-patching bias: modest but statistically significant (Wilcoxon
$p < 10^{-4}$). We report paired patching as the primary result.

Recovery is undefined when $|P_\text{ctrl} - P_\text{GP}| < 0.01$
(3/400 observations, 0.75\%); these are excluded. Denominator
diagnostics: mean~0.39, p10~0.20, confirming healthy effect sizes.

\section{BOS Token Handling}
\label{app:bos}

Pythia: BOS token exists but is not auto-prepended by the tokenizer
call. Llama-3.2: BOS is auto-prepended. Qwen3.5: no BOS token
defined. All models are evaluated with their native tokenization
behavior (no manual BOS insertion or removal).

\section{Calibration Detail}
\label{app:calib}

Contextual calibration \citep{zhao2021} on Pythia-1B: Negation 0\%
$\to$ 0\% (genuine fixation); Category 0\% $\to$ 60\% (mixed, 40pp
residual); Arithmetic 0\% $\to$ 100\% (pure label bias).

\section{Symmetry Control}
\label{app:symmetry}

Reverse direction (all-dog labels, cat test): 0\% at 70M/1B/2.8B,
10\% at 6.9B---mirroring the standard GP direction.

\section{Scale-Emergent Alternation Fixation}
\label{app:alt}

Deterministic alternating CDCDCDCD ordering causes 72--81pp gaps at
$\geq$1B Pythia (vs.\ shuffled controls). Alternation fixation is
mechanistically distinct from homogeneity fixation: the former
triggers on periodicity, the latter on unanimity.

\section{Tokenization Verification}
\label{app:tokens}

All nonsense tokens (\{foo, bar, vex, nit, orb\}) and task labels
(\{dog, cat, positive, negative, hot, cold, big, small\}) encode to
exactly one token across all four tokenizer families (GPT-NeoX,
Llama-3, Qwen3.5, Qwen3). This is verified at runtime before each
experiment via
\texttt{tokenizer.encode(f"\textvisiblespace\{label\}",
add\_special\_tokens=False)} returning a length-1 list.
Representative token IDs: Pythia \texttt{foo}$=$17374,
\texttt{bar}$=$2534, \texttt{vex}$=$49322, \texttt{nit}$=$12389,
\texttt{orb}$=$36391; Llama \texttt{foo}$=$15586, \texttt{bar}$=$3703,
\texttt{vex}$=$84265, \texttt{nit}$=$25719, \texttt{orb}$=$37466.

\section{Format Ablation}
\label{app:format}

To test whether fixation depends on a specific prompt format, we
evaluate Pythia-1B/2.8B/6.9B on the category task under five
delimiter conventions: (1)~\texttt{input: label}, (2)~\texttt{Q:
input\textbackslash nA: label}, (3)~\texttt{input -> label},
(4)~\texttt{input: x\textbackslash nlabel: y}, and
(5)~double-newline separators. Across all 15 cells (5 formats
$\times$ 3 sizes), GP accuracy remains 0\% and $P(\text{GP
label})\!>\!0.88$; fixation is invariant to delimiter choice. The
four tasks reported in the main text (colon, question-mark, and
equals conventions) further confirm cross-format robustness.

\end{document}